\documentclass[letterpaper, 10 pt, conference]{ieeeconf}  % Comment this line out if you need a4paper

\IEEEoverridecommandlockouts                              % This command is only needed if 
\usepackage[utf8x]{inputenc}
\usepackage{amsmath}
\usepackage{amssymb}
\usepackage{wrapfig}
\usepackage{graphicx}
\usepackage{algorithm2e}
\usepackage{eqparbox}
\usepackage{mdwmath}
\usepackage{mdwtab}
\usepackage{array}
\usepackage{authblk}

\usepackage{hyperref}
\usepackage{units}
\usepackage{multirow}

\usepackage{subfigure}

\usepackage{setspace}
\usepackage[usenames,pdftex]{color}
\definecolor{MyBrickRed}{cmyk}{0,0.89,0.94,0.28}
\definecolor{MyRed}{cmyk}{0,1,1,0}
\definecolor{MyGreen}{rgb}{0,1,0}

\usepackage[left=2cm,right=2cm,top=2cm,bottom=2cm]{geometry}

\usepackage[english]{babel}

\usepackage{pslatex}

\graphicspath{
	{figs/}
	{matlab/}
	{anim/}
}

\title{\LARGE \bf
Generative One-Shot Learning (GOL): A Semi-Parametric Approach to One-Shot Learning in Autonomous Vision
}

\author{Sorin M. Grigorescu% <-this % stops a space
\thanks{Sorin M. Grigorescu is with Elektrobit Automotive
		and the Faculty of Electrical Engineering and Computer Science,
        Transilvania University of Brasov, Romania.
        {\tt\small sorin.grigorescu@elektrobit.com}}%
}

\begin{document}

\maketitle
\thispagestyle{empty}
\pagestyle{empty}

\begin{abstract}

\textit{Highly Autonomous Driving} (HAD) systems rely on deep neural networks for the visual perception of the driving environment. Such networks are trained on large manually annotated databases. In this work, a semi-parametric approach to one-shot learning is proposed, with the aim of bypassing the manual annotation step required for training perceptions systems used in autonomous driving. The proposed generative framework, coined \textit{Generative One-Shot Learning} (GOL), takes as input single one-shot objects, or generic patterns, and a small set of so-called regularization samples used to drive the generative process. New synthetic data is generated as \textit{Pareto optimal solutions} from one-shot objects using a set of \textit{generalization functions} built into a \textit{generalization generator}. GOL has been evaluated on environment perception challenges encountered in autonomous vision.

\end{abstract}

%\begin{keyword}
%Object tracking \sep Active contours \sep 3D reconstruction \sep Robot vision systems
%\end{keyword}

%\end{frontmatter}

%%%%%%%%%%%%%%%%%%%%%%%%%%%%%%%%%%%%%%%%%%%%%%%%%%%%%%%%%%%%%%%%%%%%%%%%%%%%%%%%

\section{Introduction}
\label{sec:introduction}

As with many artificial intelligence systems nowadays, autonomous driving makes use primarily of supervised deep learning techniques, where object detection algorithms are trained on large manually annotated databases. The traditional processing pipeline for such methods is mainly based on two stages, where the first stage detects a potential interest region (e.g. a rectangle bounding an object of interest in an image), while the second stage classifies that region according to a set of learned object classes.

The deep learning revolution brought major improvements in the autonomous driving technology, where the environment surrounding a car can now be understood through vision and information fusion systems, as presented in the computer vision for autonomous vehicles survey from~\cite{JanaiGBG17}. Although great progress has been made in this area, fully autonomous driving in complex and arbitrarily environments is still an open challenge. In order for autonomous driving cars to be deployed on a mass scale, their artificial intelligence systems have to possess the capability to generalize and understand in real-time unpredictable scenes and situations. As stated in~\cite{JanaiGBG17}, the current perception systems used for autonomous driving are producing error rates which are not acceptable for their commercial and large scale deployment, exhibiting less robustness than a human driver.

\begin{figure}
	\centering
	\begin{center}
		\includegraphics[clip,trim=20mm 214mm 87mm 30mm, scale=0.85]{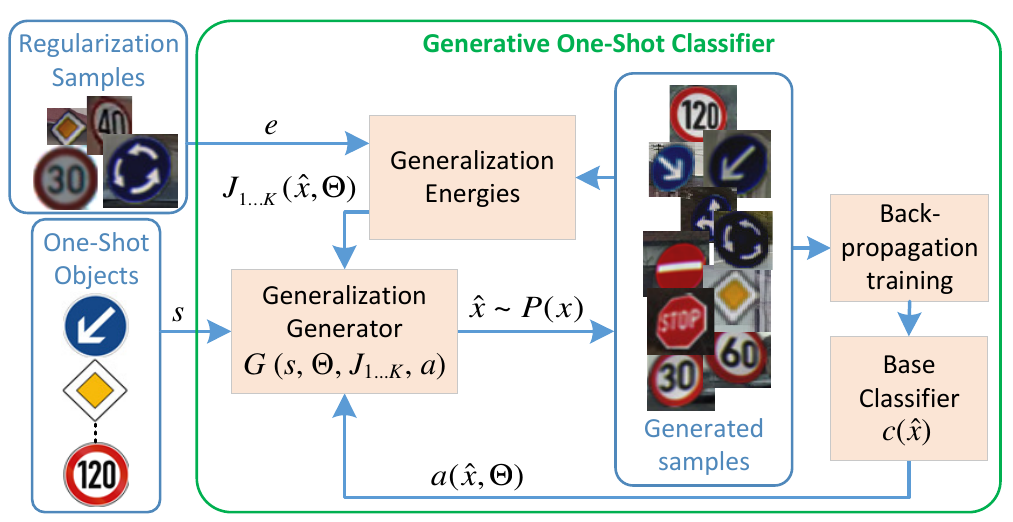}
		\caption{Summary of the Generative One-Shot Learning approach described in Section~\ref{sec:gol}. The GOL framework takes as input a set of one-shot objects $s$, along with a set of regularization samples $e$. The one-shot objects are used by the \textit{Generalization Generator} to generate synthetic samples $\hat{x} \sim P(x)$ for each object class. The synthetic samples mimic the real PDF $P(x)$. The samples generation process is governed by a set of \textit{generalization functions} $G(s, \Theta, J_{1...K}, a)$, where $\Theta$ is a vector of data generation parameters, $J_{1...K}(\hat{x}, \Theta)$ is the \textit{Generalization Energies} vector and $a(\hat{x}, \Theta)$ is the accuracy of the base classifier $c(\hat{x})$, trained via backpropagation. Generalization energies are calculated by measuring the similarity between generated data $\hat{x}$ and regularization samples $e$. The training of GOL is performed through a Pareto multi-objective optimization technique, which aims at simultaneously maximizing the generalization energies $J_{1...K}(\cdot)$ and the classification accuracy $a(\cdot)$. The final optimal base classifier $c(\hat{x}^*)$ learns data distributions from Pareto optimal synthetic samples $\hat{x}^*$.}
        \label{fig:idea}
	\end{center}
\end{figure}

In this paper, a generative semi-parametric algorithm for one-shot learning, entitled \textit{Generative One-Shot Learning} (GOL) is proposed. Its purpose is to replace supervised trained classifiers which rely on the manual collection and annotation of training data. In particular, GOL is intended to act as an AI system for environment perception in autonomous driving, which can organically adapt to new complex situations and learn from the scarce available information describing these new driving scenes. The training of GOL is performed based on conflicting objectives which imply the usage of so-called \textit{Pareto optimal solutions}, which model the fact that, for a given problem, there exists a (possible infinite) number of Pareto optimal solutions. The multi-objective optimization is also called Pareto optimization. We introduce the \textit{Pareto optimal classifier} concept for an estimator obtained from Pareto optimal solutions.

The main contributions of this paper are summarized as follows:

\begin{itemize}
	\item the GOL generative one-shot learning framework depicted in Fig.~\ref{fig:idea}, which uses a set of generalization functions to map a single training example to a base classifier;
	\item a Pareto optimization procedure for training the GOL framework through the maximization of a multi-objective function composed of generalization energies and a classification accuracy measure;
	\item the Visual GOL algorithm for environment perception in autonomous vision.
\end{itemize}

The rest of the paper is organized as follows. The GOL algorithm will be detailed in the next section, while the state of the art and the relation of GOL to other approaches will be given in the related work Section~\ref{sec:related_work}. The performance of GOL in autonomous vision tasks is presented in Section~\ref{sec:gol_for_had}. Finally, conclusions are stated in Section~\ref{sec:conclusions}.

\section{Generative One-Shot Learning}
\label{sec:gol}

\subsection{GOL Model}

The GOL approach to one-shot learning, illustrated in Fig.~\ref{fig:idea}, is based on three components:

\begin{itemize}
	\item a \textit{Generalization Generator} composed of a set of generalization functions used to generate synthetic sets of training samples from a single one-shot object instance;
	\item a \textit{Generalization Energy} measure which calculates the generalization error between synthetic generated samples and a small set of regularization samples;
	\item a \textit{Base Classifier}, implemented as a deep neural network trained on Pareto optimal synthetic samples.
\end{itemize}

The proposed model architecture aims at obtaining a stable base classifier $c(\hat{x})$ trained on a set of synthetic samples $\hat{x} \sim P(\hat{x})$ generated from one-shot instances of objects of interest. The generative nature of the model tries to capture the \textit{Probability Density Function} (PDF) $P(\hat{x})$ as close as possible to the real data PDF $P(x)$, from which real samples $x$ are drawn. Each one-shot object $s_i$ is stored in the one-shot objects set $s$:

\begin{equation}
	s = \{ (s_i, y_i) \}_{i=1}^{K},
	\label{eq:one_shot_objects_set}
\end{equation}

\noindent where $s_i$ is a one-shot instance of an object with label $y_i$, meaning that the size of $s$ corresponds to the $K$ number of object classes which the base classifier $c(\hat{x})$ will learn. Each synthetic sample from $\hat{x}$ will have a corresponding label $y$:

\begin{equation}
	\hat{x} = \{ (\hat{x}_i, y_i) \}_{i=1}^{m},
	\label{eq:one_shot_labels_set}
\end{equation}

\noindent where $m >> K$. Typically, the number of generated synthetic samples $m$ has a very large value, depending on the number of classes we want to classify, as well as on the size of the feature vector describing an object.

The goal of the generalization generator is to generate a set $\hat{x}$ which describes as close as possible the real probability density $P(x)$ from which real-objects are drawn: $x \sim P(x)$. The synthetic descriptions $\hat{x}$ are obtained by applying a series of generalization functions from the generalization generator $G(\Theta, \cdot)$:

\begin{equation}
	G(\Theta, \cdot) = \begin{bmatrix} g_1(\theta_1, \cdot), g_2(\theta_2, \cdot), ... , g_i(\theta_i, \cdot), ... , g_n(\theta_n, \cdot) \end{bmatrix},
	\label{eq:gol_multiobjective_optimization}
\end{equation}

\noindent where $G(\Theta, \cdot)$ is a family of $n$ functions $g(\theta, \cdot)$ which, applied iteratively on a one-shot instance $s_i$, will generate  a set of synthetic descriptions $\hat{x}$. $\theta_i$ holds the parameters of function $g_i(\theta_i, \cdot)$. For the sake of clarity, although $\theta$ might have variable size, depending on the number of parameters a function $g(\theta, \cdot)$ takes, we will consider $\theta$ to be a scalar. The parameters of all functions $g_i(\theta_i, \cdot)$ from the generalization generator, where $i = 1, ..., n$, are stored in the parameters vector $\Theta$:

\begin{equation}
	\Theta = \begin{bmatrix} \theta_1, \theta_2, ... , \theta_i, ... , \theta_n \end{bmatrix},
	\label{eq:gol_parameters_matrix}
\end{equation}

\noindent As it is shown in the next subsection, the Pareto optimization procedure of GOL learns a collection $\Theta^* \in \mathbb{R}^{n \times l}$ of optimal generalization functions parameters:

\begin{equation}
	\Theta^* = \begin{bmatrix} \Theta_1^*, \Theta_2^*, ... , \Theta_l^*, \end{bmatrix}^T,
	\label{eq:optimal_thetas}
\end{equation}

\noindent where $l$ is the number of Pareto optimal solutions obtained via GOL training. Each $\Theta_i^*$ will generate an optimal synthetic samples set $\hat{x}^*$ describing a one-shot object.

 %% IMPORTANT
 %% \Thetas nu au solutii fixe optime, ci solutii pareto, facand astfel posibila generarea de mai multi parametrii Theta, pentru fiecare valoare a lor generandu-se un set de date artificiale de antrenare

In order to improve the convergence of the algorithm, the concept of \textit{regularization samples}, or \textit{experiences} $e \sim P(x)$ is introduced. In comparison to the number of generated samples, the number of regularization samples $e$ is much smaller: $\dim{(e)} << m$. The goal of the regularization samples is to ensure that $\hat{x}$ is as close as possible to the real data distribution $P(x)$. The base classifier $c(\hat{x})$ cannot be trained solely on regularization samples due to their small number, which would, in turn, create a very large generalization error in $c(\hat{x})$. The probability of class $k$ for example $i$ is computed using standard softmax function.

At its core, GOL models a mapping between one-shot objects $s$ and a base classifier, $s \rightarrow c(\cdot)$, based on calculated synthetic samples $\hat{x}$. The probabilistic model of GOL can be described as a PDF which generates $\hat{x}$ from the Generalization Generator $G(\cdot)$, given a set of one-shot objects and a small set of regularization samples:

\begin{equation}
	P(\hat{x} | s, e, \Theta) = \mathcal{N} (e; G(\Theta, s)).
	\label{eq:model_1}
\end{equation}

%\begin{equation}
%	P(\hat{x} | S, E, \Theta) = \mathcal{L} ( G(s, \Theta) )
%	\label{eq:model_2}
%\end{equation}

%\noindent where $\mathcal{L}(\cdot)$ denotes the log-likelihood.

The semi-parametric nature of GOL lies in the way in which the model in Eq.~\ref{eq:model_1} is learned. The algorithm determines its parameters as a set of Pareto optimal solutions by iteratively generating synthetic samples, thus driving the classification accuracy and the values of the generalization energies. As described in the next section, a generalization energy is a distance metric specifying how well GOL generalizes classification through its base classifier.

\subsection{Training Strategy}
\label{sec:gol_training}

GOL is intended to generalize on unseen data, while increasing the classification accuracy on synthetic data as well as possible. The training of GOL implies the learning of a set of optimal parameters $\Theta^*$ which maximize the generalization energies and classification accuracy. A Pareto optimization procedure is proposed for training the GOL system. The multi-objective nature of the training is due to the multitude of functional values that have to be optimized and also because there exists not only one instance of optimal GOL parameters, but a number of Pareto optimal solutions defined as $\Theta^*$ in Eq.~\ref{eq:optimal_thetas}. Due to the very scarce information on the real data's PDF, a single objective training approach would create a base classifier which would overfit. In order to distinguish between the overall multi-objective training of GOL and the training of the base classifier on generated synthetic samples, we will refer to a Pareto optimization iteration as a training episode, while each learning iteration of the base classifier will be referred to as a training epoch. One GOL training episode contains several hundred training epochs.

\subsubsection{Introduction to Pareto Optimization}

GOL training aims to simultaneously maximize the generalization energies and classification accuracy in a Pareto optimization manner. The optimization procedure does not search for a fixed size, or even finite, set of model parameters, but for a Pareto optimal solutions set.

In the following, $\Theta = \begin{bmatrix} \theta_1, \theta_2, ..., \theta_i, ..., \theta_n, \end{bmatrix}^T$ from Eq.~\ref{eq:gol_parameters_matrix} will be called a solution vector, composed of $n$ decision variables $\theta_i$, with $i = 1, ..., n$ and $\Theta \in \mathbb{R}^n$.

In \textit{scalarized} multi-objective learning, the different measures are aggregated into a single scalar cost function $F$, which is to be either minimized or maximized:

\begin{equation}
	\begin{aligned}
		\text{minimize / maximize } & F( J_w (\Theta), \lambda ), && w = 1, 2, ..., W \\
		\text{such that } & \Theta \in \Theta_{\lambda}, &&
	\end{aligned}
	\label{eq:scalarized_optimization}
\end{equation}

\noindent where $F: \mathbb{R}^{W+1} \mapsto \mathbb{R}$ is a function which aggregates all $W$ cost functions based on the scalarization values given by vector $\lambda$, such that $\Theta \in \Theta_{\lambda}$:

\begin{equation}
	F( J_w (\Theta), \lambda ) = \min / \max \sum_{i=1}^W \lambda_i J_i (\Theta)
	\label{eq:linear_scalarized_optimization}
\end{equation}

The main issue with scalarized optimization is the a-priori knowledge required in setting up the weighting vector $\lambda$. This means that the importance of each cost function in Eq.~\ref{eq:linear_scalarized_optimization} has to be manually defined in advance, thus possibly driving the optimization to local minimas or maximas.

Generalizing from Eq.~\ref{eq:scalarized_optimization}, the multi-objective optimization problem takes into account a number of functional values which have to be either minimized or maximized, subject to a number of possible constraints that any feasible solution must satisfy~\cite{Deb11}:

\begin{equation}
	\begin{aligned}
		\text{minimize / maximize } & J_w(\Theta), && w = 1, 2, ..., W \\
		\text{subject to } & r_v (\Theta) \geq 0, && v = 1, 2, ..., V \\
		& h_q (\Theta) = 0, && q = 1, 2, ..., Q \\
		& \theta_i^{(L)} \leq \theta_i \leq \theta_i^{(U)}, && i = 1, 2, ..., n
	\end{aligned}
	\label{eq:multiobjective_optimization}
\end{equation}

\noindent where $J_w(\Theta)$ is the $w$-th objective function, with $w=1,~..., W$, and $W \geq 2$ is the number of objectives. $r_v (\Theta)$ and $h_q (\Theta)$ are constraints set on the optimization process. $\theta_i^{(L)}$ and $\theta_i^{(U)}$ are lower, respectively upper, variable constraint bounds set on each decision variable.

\begin{figure}
	\centering
	\begin{center}
		\includegraphics[scale=0.9]{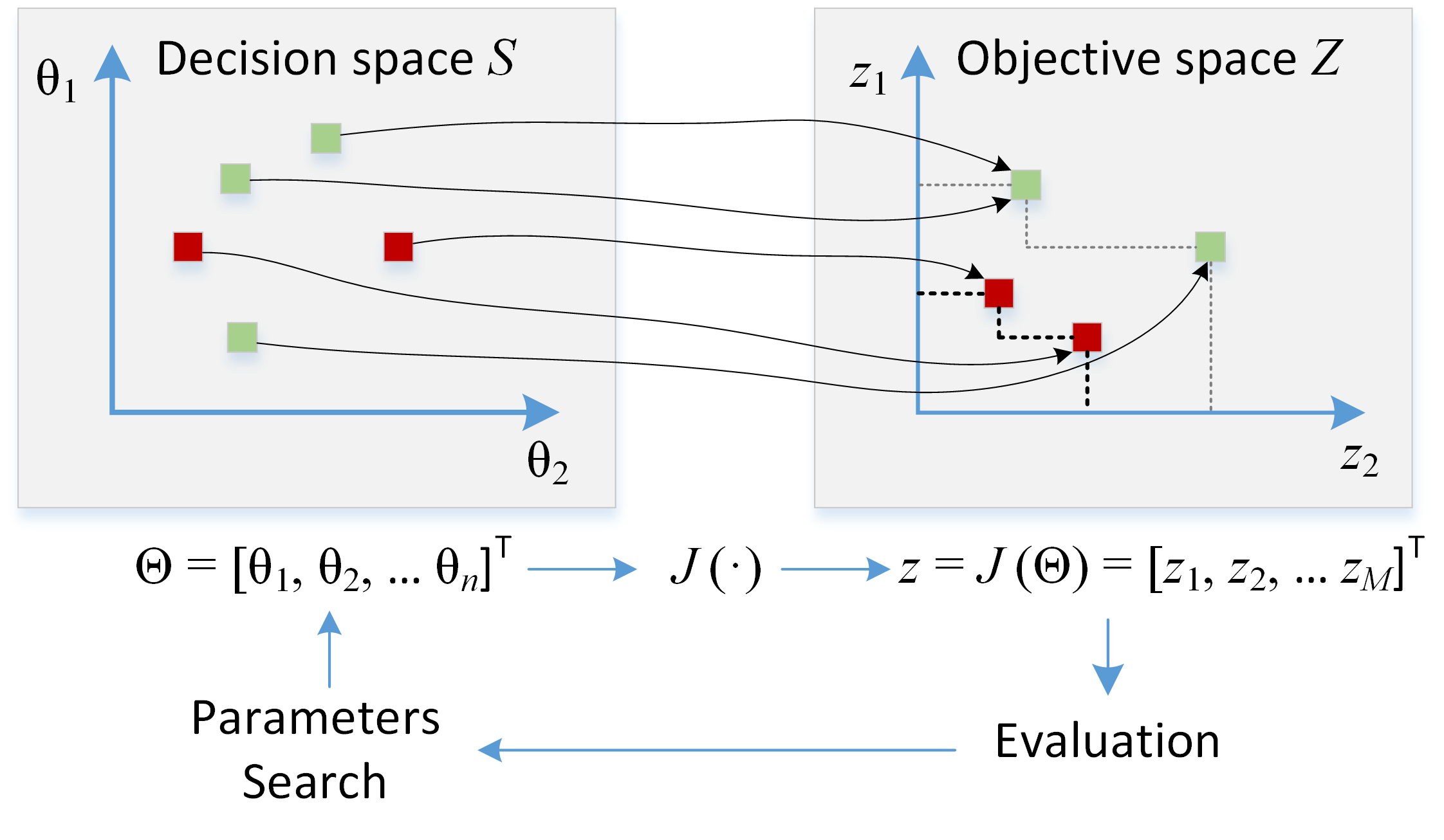}
		\caption{Mapping of solution vectors $\Theta$ from the \textit{decision space} $S$ to \textit{objective space} $Z$. Each solution $\Theta$ in decision space corresponds to a coordinate $z$ in objective space. The red marked coordinates are the set of Pareto optimal solutions $\Theta^*$ for a multi-objective minimization problem, located on the Pareto front drawn with thick black line.}
        \label{fig:pareto_optimization_problem}
	\end{center}
\end{figure}

The solutions satisfying $r_v (\Theta)$ and $h_q (\Theta)$ and the variable bounds form the so-called \textit{feasible decision variable space} $S \in \mathbb{R}^n$, or simply \textit{decision space}. A core difference between single and multi-objective optimization is that, in the latter case, the objective functions make up a $W$-dimensional space entitled \textit{objective space} $Z \in \mathbb{R}^W$. A visual illustration of a 2D decision and objective space is visible in Fig.~\ref{fig:pareto_optimization_problem}. For each solution $\Theta$ in the decision variable space, there exists a coordinate $z \in \mathbb{R}^M$ in the objective space:

\begin{equation}
	z = J (\Theta) = \begin{bmatrix} z_1, z_2, ..., z_M, \end{bmatrix}^T
	\label{eq:coordinate_objective_space}
\end{equation}

A solution is a variable vector in decision space, with a coordinate $z$ as a corresponding objective vector. In Pareto optimization there exists a set of optimal solutions $\Theta^*$, none of them usually minimizing, or maximizing, all objective functions simultaneously. Optimal solutions are called \textit{Pareto optimal}, meaning that they cannot be improved in any of the objectives without degrading at least one objective. A feasible solution $\Theta^{(1)}$ is said to Pareto dominate another solution $\Theta^{(2)}$ if:

\begin{enumerate}
	\item $J_i (\Theta^{(1)}) \leq J_i (\Theta^{(2)})$ for all $i \in \{1, 2, ..., W\}$ and
	\item $J_j (\Theta^{(1)}) < J_j (\Theta^{(2)})$ for at least one index $j \in \{1, 2, ..., W\}$.
\end{enumerate}

A solution $\Theta^*$ is called Pareto optimal if there is no other solution that dominates it. The set of Pareto optimal solutions is entitled \textit{Pareto boundary}, or \textit{Pareto front}. As shown in Fig.~\ref{fig:pareto_optimization_problem}, the search for optimal solutions is performed in decision space, while their evaluation takes place in objective space. State of the art methods for calculating Pareto boundaries are based on evolutionary computing, which is out of the scope of this paper. Further details on Pareto optimization using evolutionary methods can be found in~\cite{Deb11}, while a few case studies in the field of machine learning are available in~\cite{Jin2008}.

\subsubsection{GOL Training via Pareto Optimization}

Within GOL, the optimization objective is to find the set of Pareto optimal solutions $\Theta^*$ which can be used to generate synthetic data from one-shot objects. The training is made in variable $t \in [1, T]$ episodes, making it possible to constantly improve the generalization capabilities of the $G(\cdot)$ functions, once new regularization samples are available. For the sake of simplicity, we will write a vector of generalization energies $[J_1(\cdot), J_2(\cdot), ..., J_K(\cdot)]$ as $[J_{1...K}(\cdot)]$.

Following the Pareto optimization framework from Eq.~\ref{eq:multiobjective_optimization}, the GOL Pareto optimization problem can be formulated as:

\begin{equation}
	\begin{aligned}
		\text{maximize } && \begin{bmatrix} J_1(\hat{x}, \Theta), J_2(\hat{x}, \Theta), ..., J_k(\hat{x}, \Theta), a(\hat{x}, \Theta) \end{bmatrix} && \\
		\text{such that } && \theta_i^{(L)} \leq \theta_i \leq \theta_i^{(U)}, \text{ }\text{ }\text{ }\text{ }\text{ }\text{ }\text{ }\text{ }\text{ }\text{ }\text{ }\text{ }\text{ }\text{ } &&
	\end{aligned}
	\label{eq:gol_multiobjective_optimization}
\end{equation}

\noindent where $J_1(\hat{x}, \Theta)$ to $J_K(\hat{x}, \Theta)$ and $a(\hat{x}, \Theta)$ are the generalization energies and classification accuracy cost functions, respectively. $K$ is the number of input one-shot objects and $i = 1, 2, ..., n$. The proposed GOL training procedure is not driven by the inequality and equality constraints from Eq.~\ref{eq:multiobjective_optimization}, but bounded by the lower and upper variable constraint bounds $\theta_i^{(L)}$ and $\theta_i^{(U)}$.
% The negative sign associated with the classification accuracy measure is intended to convert the maximization of the accuracy to a minimization problem.

The generalization energies $J_{1...K}(\cdot)$ are a vector of real-valued functions defined as log-likelihood similarity metrics between the regularization samples $e$ and the synthetic data $\hat{x}$, for each object class $k$:

\begin{equation}
	J_k(\Theta, \hat{x}(t)) = \mathbb{E}_{\hat{x} \sim P(\hat{x}), e \sim P(x)} \log p_{model} (k | \hat{x})
	\label{eq:generalization_loss}
\end{equation}

\noindent where $t$ is a training episode. Eq.~\ref{eq:generalization_loss} expresses the probability of the synthetic data to be drawn from the regularization samples. $J_k(\Theta, \hat{x}(t))$ varies depending on the model and can be described as the following error cost:

\begin{equation}
	J_k(\Theta, \hat{x}(t)) = \mathbb{E}_{\hat{x} \sim P(\hat{x}), e \sim P(x)} ||e - \hat{x}(t)||_q + const
	\label{eq:generalization_loss_unfolded}
\end{equation}

\noindent where $|| \cdot ||$ is a norm function, $q \geq 1$ represents the $q$-norm (also called ${\displaystyle \ell _{q}}$-norm) and $const$ is a random variable chosen based on the variance of the Gaussian distribution implied by the GOL model in Eq.~\ref{eq:model_1}.

$a(\Theta, \hat{x}(t)) \in [0, 1]$ is the classification accuracy, which has a probability value of $1$ if all synthetic samples have been classified correctly and $0$ if none is correctly classified:

\begin{equation}
	a(\Theta, \hat{x}(t)) = 	\frac{ \sum_{i=1}^m 1\{ \arg \max_k p(c (\hat{x})) = y_i \} } {m}
	\label{eq:classification_accuracy}
\end{equation}

\noindent where $\arg \max p(c (\hat{x}))$ returns the predicted class of the classifier, $p(\cdot)$ is the softmax function and $m$ represents the number of synthetic samples. $1\{ \cdot \}$ is an indicator function which returns $1$ if the expression evaluated within the brackets is true and $0$ otherwise.

With each training episode, the variance of the GOL's $\Theta$ parameters is increased by an additive factor $\Delta = [\delta_1, \delta_2, ..., \delta_n]$, where the number of elements in $\Delta$ is equal to the number of generalization parameters. At the end of an episode, pairs of \textit{solutions} - \textit{objective functional values} are stored in a container variable:

\begin{equation}
	\Xi \leftarrow \{ \Theta, J_1(\Theta, \hat{x}(t)), J_2(\Theta, \hat{x}(t)), ..., J_K(\Theta, \hat{x}(t)), a(\Theta, \hat{x}(t)) \}
	\label{eq:set_solutions_objectives}
\end{equation}

The Pareto optimal solutions $\Theta^*$, are determined from the $\Xi$ set, based on the evolutionary computation approach described in~\cite{Deb11}. We call the final $c(\hat{x}^*)$ estimator a Pareto optimal classifier, obtained by regenerating synthetic data for all Pareto optimal parameters:

\begin{equation}
	\hat{x}^* \sim P(G(\Theta^*, s))
	\label{eq:pareto_optimal_solutions}
\end{equation}

\section{Related Work}
\label{sec:related_work}

Similar approaches to GOL can be classified into \textit{zero-shot learning}, focused on learning based on attribute-level descriptions of object classes, \textit{one-shot learning}, aiming to train classifiers based on a The challenge of one-shot learning has been treated broadly in the work of Fei-Fei~\cite{fei_fei_pami2006, fei2006knowledge} for the case of visual object recognition. Their approach is based on a Bayesian framework, named "Bayesian One-Shot", which learns new object categories based on knowledge, or priors, acquired from previously learned categories. In comparison to Bayesian One-Shot, GOL does not rely on large training databases priors for learning object classes. Also, the "Bayesian One-Shot" framework was built for the explicit purpose of visual recognition, as opposed to GOL, which is a general one-shot learning framework that can be adapted to completely new problems solely by changing the structure of the Generalization Generator's functions $G(\Theta, \cdot)$. 

Hand-crafted attribute descriptions of novel categories have been used in~\cite{Farhadi2010, Lampert2014, Romera-Paredes2017} for training classifiers for image recognition. Methods based on metric learning for automatic feature representation of object classes have been reported in~\cite{Fink2004, Schroff2015, Taigman2015}. Apart from the strictly image domain application of these algorithms, one of the main constraints of these methods is that the features of the objects of the same class should be clustered together.

OpenAI published an interesting robotics application of one-shot learning, described in~\cite{DuanASHSSAZ17}. Their algorithm, named \textit{one-shot imitation learning}, uses a meta-learning framework for training robotic systems in performing certain tasks based only on a couple of demonstrations. Also, the large amount of data required by Deep Reinforcement Learning systems has been approached in~\cite{WangKTSLMBKB16} through the introduction of their \textit{deep meta-reinforcement learning} algorithm.

\textit{Siamese Neural Networks} have been investigated for the purpose of one-shot character recognition in images in the work of Koch et. al.~\cite{Koch2015SiameseNN}. Such networks have a structure which allows them to naturally rank similarity between inputs.  Deep reinforcement learning is used in~\cite{WoodwardF17} for active one-shot learning, where a recurrent neural network has been trained to estimate an action-value function which indicates what examples from a stream of images are worth labeling. A \textit{low-shot} visual object recognition method was proposed by Facebook AI Research in~\cite{HariharanG16}, where the focus is on datasets with object categories that have high intra-class variation. In comparison to the GOL approach, the siamese networks~\cite{Koch2015SiameseNN}, as well as the low-shot algorithm from~\cite{HariharanG16}, rely on extensively pre-trained models which must discriminate between class-identity of image pairs. The basic assumptions encountered in both papers is that a deep neural network which performs well when tested against images belonging to the learned categories, should generalize well to one-shot learning classification when an image belonging to a new category is presented. This statement is not entirely true, since there exist very similar objects in appearance, but actually belonging to different classes. As it will also be shown in the next subsection, the necessity of a pre-trained classifier on which the learning of new object classes leverages on is the core difference between GOL and deep one-shot learning. Also, in both methods, a high intra-class variance between object categories is a necessary prior, the algorithms accuracies decreasing proportional to the decrease of intra-class variance.

The output of the GOL's training loop can be compared to a form of automatic data augmentation. In~\cite{Krizhevsky_2012}, two data augmentation methods are used for increasing the size of the training database. The first form of data augmentation generates images by translations and reflections, while the second form uses PCA for altering the pixels values. Within GOL, data augmentation is a side effect produced by its training loop, where the augmentation parameters are automatically determined by the optimization procedure. Common data augmentation approaches, like the ones in~\cite{Krizhevsky_2012}, use fixed, manually determined, parameters for generating new samples, without taking into consideration any quality measurement regarding the nature of the synthetic data.

Google's DeepMind approached the one-shot learning problem through the usage of \textit{Memory Augmented}~\cite{SantoroBBWL16} and \textit{Matching Neural Networks}~\cite{VinyalsBLKW16}. Such networks allow more expressive models based on the introduction of "content" based attention and "computer-like" architectures such as the Neural Turing Machine~\cite{GravesWD14}, or Memory Networks~\cite{WestonCB14}. The DeepMind research group that introduced matching networks to one-shot learning, also reported the usage of memory-augmented neural networks for attacking the same problem~\cite{SantoroBBWL16}. For this task, an external-memory equipped network has been used. In comparison to GOL, matching and memory-augmented networks require not a single one-shot example for learning, but an external pre-trained neural net, on which later learning is leveraged on, thus failing to avoid the standard supervised training of their systems from large labeled training databases. As in~\cite{fei_fei_pami2006}, previously supervised learned classes are required as a base for one-shot learning with matching and memory-augmented networks, making the presented solutions to fall more into the area of transfer learning than one-shot learning. Also, for incorporating newly unseen information, these networks require a support set of labeled examples. Even if small, the labeled support set is required for training, as opposed to GOL, where the regularization samples are not strictly necessary, their purpose being only to drive the learning process, if such regularization samples exist.

One of the most influential work on GOL is the research on \textit{Generative Adversarial Nets} (GAN)~\cite{Goodfellow2014}. The major difference between GOL and the adversarial nets is that, within GOL, the generation of synthetic information is performed based on the generalization functions which generate new data from a one-shot object, thus making GOL a pure one-shot learning framework. 

The \textit{Deep Recurrent Attentive Writer} (DRAW)~\cite{icml2015_gregor15} is a variational auto-encoder composed of a pair of recurrent neural networks, one functioning as an encoder, while the other one acts as a decoder that reconstitutes images after receiving codes. More focused on the problem of one-shot learning, the work in~\cite{Rezende2016OneShotGI} focuses on the introduction of \textit{sequential generative models} which, once trained, can generate new object samples. If the objective of these methods is the generation of new synthetic samples, the GOL approach differs in the sense that GOL generates samples in order to train its classifier and not the other way around. Hence, in GOL we are obtaining a large bundle of synthetic data, where the inner class boundaries are calculated using the generalization energies and the classification accuracy. Also, as deep learning techniques, GANs~\cite{Goodfellow2014}, DRAW~\cite{icml2015_gregor15} and the sequential generative models~\cite{Rezende2016OneShotGI} require large labeled training databases for training. In GOL, the learning starts from generated synthetic data which is very similar to the one-shot input object. As the training progresses, the new synthetic data is generalized into the neural classifier until the generalization loss and classification accuracy constraints are broken. Due to this different generative starting point in training, the convergence of GOL to an optimal classifier is much faster than in, for example, GANs.

\begin{figure*}
	\centering
	\begin{center}
		\includegraphics[scale=0.38]{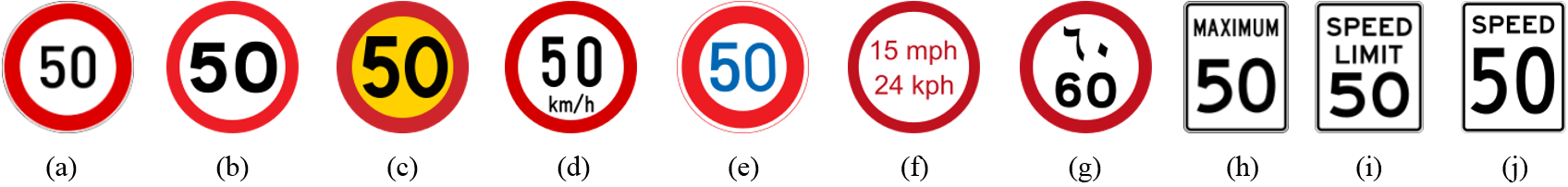}
		\caption{Traffic sign templates for speed limits in different countries [source \url{www.wikipedia.org}]. (a) Vienna convention. (b) United Kingdom. (c) Alternative Vienna convention. (d) Ireland. (e) Japan. (f) Samoa. (g) United Arab Emirates and Saudi Arabia. (h) Canada. (i) United States. (j) United States (Oregon variant).}
        \label{fig:ts_50}
	\end{center}
\end{figure*}

\begin{figure*}
	\centering
	\begin{center}
		\includegraphics[scale=0.88]{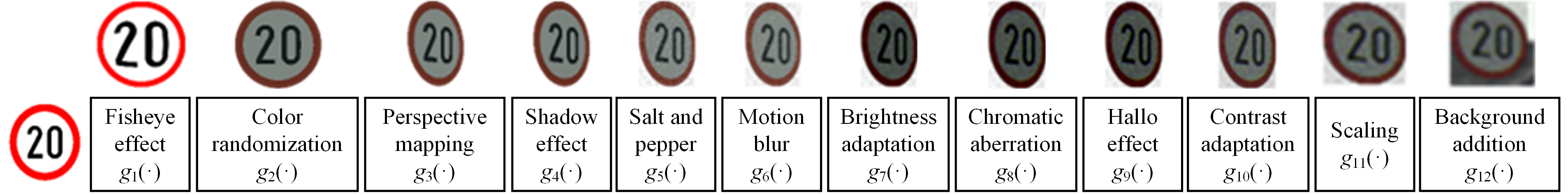}
		\caption{Coupled stages of the Generalization Generator functions in the image domain. The one-shot object representing the speed limit sign $20$ kmh is recursively passed through the generalization functions, with different $\Theta$ parameters at each episode.}
        \label{fig:image_based_gol}
	\end{center}
\end{figure*}

\section{Performance of Visual Object Recognition of GOL in Autonomous Vision}
\label{sec:gol_for_had}

\subsection{Generative One-Shot Learning for Visual Perception}
\label{sec:visual_gol}

With the advent of deep learning, the performance of visual recognition systems surpassed in some use cases even the recognition capabilities of humans~\cite{Krizhevsky_2012}. Visual object recognition nowadays relies mainly on large manually labeled training databases with which deep convolutional neural networks are trained in a supervised manner. The acquisition and manual labeling of such databases is a tedious and prone to error task, especially when many object categories are involved, each category requiring thousands of manually labeled image samples. This is a common case in autonomous vision~\cite{JanaiGBG17}, as shown in the traffic signs example from Fig.~\ref{fig:ts_50}, where speed limits have different pictograms, depending on their country of origin. In order for traffic signs recognition methods to perform well, specific training data composed of thousands, or even millions, of manually labeled training samples is required for each country. The manual acquisition of a sufficient amount of labeled traffic signs is a time consuming process, due to different positions, illuminations or weather conditions that have to be taken into consideration.

In order to overcome the supervised training challenge based on manually annotated data, the \textit{Visual GOL} algorithm has been implemented in autonomous vision for the recognition of objects of interest on the road, objects such as traffic signs. Provided a set of one-shot road sign templates $S$ and a small set of regularization samples $e$, Visual GOL is able to obtain a deep convolutional neural network $c(\hat{x})$ which can reliably recognize road sign categories in real-world images. The adaptation of GOL to the image domain implies the definition of the Generalization Generator and the structure of the generalization energies.

The Generalization Generator of Visual GOL is composed of a set of $12$ image based generalization functions $G_{1...12}(\Theta, \cdot)$, defined as in Eq.~\ref{eq:gol_multiobjective_optimization}. Once applied on one-shot objects, the $G_{1...12}(\Theta, \cdot)$ functions will generate synthetic data. Fig.~\ref{fig:image_based_gol} shows coupled stages of these function defined in the image domain. The stages are realized as an ordered list of different image alteration methods, applied on the input one-shot object. Each of the $12$ $g_i(\Theta, \cdot)$ functions is described in the following paragraphs.

The one-shot input data for the Generalization Generator in Fig.~\ref{fig:image_based_gol} is a list of images, each of them representing an object category that has to be learned by Visual GOL.

The generalization energies $J_{1...K}(\Theta, \hat{x}(t))$ are defined as Bhattacharyya distances between the synthetic data and the regularization samples:

\begin{equation}
	J_k(\Theta, \hat{x}(t)) = \mathlarger{\mathlarger{\sum}}_{i=1}^{q} \mathlarger{\mathlarger{\sum}}_{j=1}^{m} \sqrt{ 1 - \frac{1}{\sqrt{\bar{s}^{(i)} \cdot \hat{\bar{x}}^{(i,j)} \cdot N^2}} \cdot \sum_{v=0}^{N} \sqrt{s^{(i)}_v \cdot \hat{x}^{(i,j)}_v} },
	\label{eq:bhattacharya_metric}
\end{equation}

\noindent where $ k = 1, ..., K$, $q$ is the number of regularization samples $e_s \sim P(x)$ of object class $s$, $m$ is the number of generated synthetic samples and $N$ is the feature vector's length. $\hat{x}^{(i,j)}$ denotes the $j$-th sample in the dataset corresponding to the $i$-th object $s^{(i)}$ of the same class $K$. In other words, the comparison between the synthetic and the regularization samples is made only between objects of the same class. $\bar{s}$ and $\bar{x}$ are the mean vectors of the two sample sets, respectively.

\subsection{Performance Evaluation on the German Traffic Sign Recognition Benchmark}

To the best of the author's knowledge, this paper is the first one to consider the application of one-shot learning to the traffic signs recognition challenge.

Road signs recognition is a mandatory functionality of highly autonomous driving cars. The \textit{German Traffic Sign Recognition Benchmark} (GTSRB)~\cite{Stallkamp2012} is a publicly available collection of $51.840$ images of German road signs, organized in $43$ classes. Since 2011 it has been a benchmark for testing traffic sign recognition systems, considering also as baseline the performance of human subjects at recognizing signs.

\begin{table*}
	\centering
	\begin{tabular}{lcccccc}
		\hline
		& Speed limits & Other prohibitions & Derestriction & Mandatory & Danger & Unique \\
		\hline
		Human (best individual)~\cite{Stallkamp2012} & 98.32 & 99.87 & 98.89 & 100.00 & 99.21 & 100.00 \\
		Human (average)~\cite{Stallkamp2012} & 97.63 & 99.93 & 98.89 & 99.72 & 98.67 & 100.00 \\
		Committee of CNNs~\cite{Ciresan2011} & 99.47 & 99.93 & 99.72 & 99.89 & 99.07 & 99.22 \\
		Multi-Scale CNN~\cite{Sermanet2011} & 98.61 & 99.87 & 94.44 & 97.18 & 98.03 & 98.63 \\
		Random Forests~\cite{Zaklouta2011} & 95.95 & 99.13 & 87.50 & 99.27 & 92.08 & 98.73 \\
		LDA (baseline)~\cite{hastie_09} & 95.37 & 96.80 & 85.83 & 97.18 & 93.73 & 98.63 \\
		\hline
		GOL [LeNet] & 98.79 & 99.47 & 96.61 & 97.43 & 98.48 & 99.16 \\
		GOL [AlexNet] & 99.49 & 99.93 & 99.74 & 99.89 & 99.53 & 99.28 \\
		GOL [GoogleNet] & 99.51 & 99.95 & 99.74 & 99.90 & 99.86 & 99.45\\
		\hline
	\end{tabular}
	\caption{Recognition accuracy results, in percentages, of state-of-the-art supervised object recognition systems against visual GOL trained on all training data from GTSRB.}
	\label{tab:ts_benchmark}
\end{table*}

The dataset has been created from 10 hours of video recorded while driving on different road types. The images are stored in raw Bayer-pattern format and have a resolution of $1360 \times 1024$ pixels. In each image, the size of the traffic sign varies from $15 \times 15$ to $222 \time 193$ pixels. The signs are annotated using a Region of Interest (ROI) which has a $10$\% margin of approximately $5$ pixels. The resulted image collection has been divided into a \textit{full training set} and a \textit{test set}. The full training set is ordered by class and has been further splitted into training and validation sets. On top of the collected raw color images, the database also provides features computed from raw data, such as \textit{Histogram of Oriented Gradients}, \textit{HAAR-like} features and \textit{color histograms}.

The supervised learning algorithms used for comparison are two types of \textit{Convolutional Neural Networks} (CNN), a \textit{Random Forest} and a \textit{Linear Discriminant Analysis} (LDA) learning algorithm, the latter two being trained on Histogram of Oriented Gradients features. LDA~\cite{hastie_09} is considered as a baseline classifier, assuming that the class densities are multi-variate Gaussians with a common covariance matrix. The Random Forest classifier~\cite{Zaklouta2011} is composed of $500$ trees, trained also on Histogram of Oriented Gradients features. The neural network based classifiers are a committee of CNNs forming a multi-column deep neural network~\cite{Ciresan2011} and a multi-scale CNN~\cite{Sermanet2011}.

Visual GOL has been trained on 43 road type object classes belonging to $6$ road sign categories, as described in~\cite{Stallkamp2012}. The 43 classes are composed of the following signs: $8$ speed limits ranging from $20$ kmh to $120$ kmh, $4$ prohibitory signs, $4$ derestriction signs, $8$ mandatory signs, $15$ danger signs and $4$ uniques signs. Similar to the approach used when testing on Omniglot, during the training of GOL each traffic sign was used as a one-shot object, while the other signs of the same class were taken as regularization samples. A bundle of synthetic samples are generated while training. As in the pedagogical example from Section~\ref{sec:gol_training}, a generalization energy is calculated for each road sign class.

\begin{figure}
	\centering
	\begin{center}
		\includegraphics[scale=0.7]{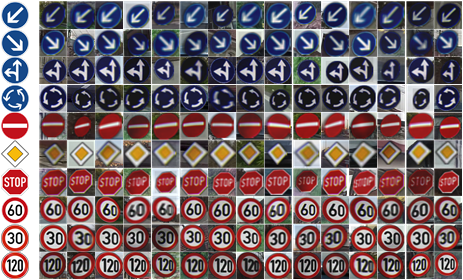}
		\caption{Synthetic pareto optimal solutions samples generated during the training of visual GOL for autonomous vision. Column 1: one-shot template object. Columns 2-16: synthetic samples generated at different training stages. [The synthetic samples generation for a couple of road signs can be viewed in the videos accompanying this paper].}
        \label{fig:generated_ts_images}
	\end{center}
\end{figure}

Apart from the synthetic data obtained based on each image in GTSRB, an additional artificial set was generated from template images of road signs, as the ones shown in the first column of Fig.~\ref{fig:generated_ts_images}. The rest of the columns in Fig.~\ref{fig:generated_ts_images} show synthetic images derived solely from the template road signs.

Experimental results obtained on the GTRSB database are presented in Table~\ref{tab:ts_benchmark}. Due to the generalization effect obtained through its generative process, Visual GOL is able to surpass the other methods when the base classifier is either AlexNet or GoogleNet. This result is mainly achieved by incorporating variance in the training data through the algorithm's generative process, as well as based on the information gathered from the template objects.

Another experimental trial covered here is one-shot object recognition of road signs when training only on template objects, such as the ones visible in the first column of Fig.~\ref{fig:generated_ts_images}. For this case, a number of $43$ road sign templates are used as training data, while $100$ images from GTSRB act as regularization samples for each template object of the same class. Recognition accuracy results are presented in Table~\ref{tab:ts_oneshot_benchmark}. Although the precision is decreased when compared to the values in Table~\ref{tab:ts_benchmark}, it is important to note that a fairly high degree of accuracy is reached, even if the method has been trained solely on template objects. This highlights the potential of GOL to overcome the rigidity of traditional supervised training approaches which leverage on large training databases.

\begin{table*}
	\centering
	\begin{tabular}{lcccccc}
		\hline
		& Speed limits & Other prohibitions & Derestriction & Mandatory & Danger & Unique \\
		\hline
		GOL [LeNet] & 82.27 & 85.21 & 83.37 & 81.64 & 83.11 & 79.80 \\
		GOL [AlexNet] & 91.16 & 90.95 & 84.38 & 89.72 & 88.22 & 87.18 \\
		GOL [GoogleNet] & 92.52 & 93.20 & 87.96 & 90.38 & 90.04 & 88.31 \\
		\hline
	\end{tabular}
	\caption{Recognition accuracy results on GTSRB, in percentages, for visual GOL trained with one-shot traffic sign templates.}
	\label{tab:ts_oneshot_benchmark}
\end{table*}

\section{Conclusions}
\label{sec:conclusions}

A current challenge in Artificial Intelligence is the development of methods that can learn and generalize well from single training examples, a problem also known as one-shot learning. In this paper, the \textit{Generative One-Shot Learning} approach has been proposed, which combines a generalization generator and a multi-objective optimization framework, that maximizes a set of generalization functions along with a classification accuracy, with the purpose of obtaining an optimal one-shot classifier. As opposed to other state-of-the-art one-shot learning methods, GOL does not require large training databases, but single object, or pattern instances. During its training, the algorithm generates various synthetic samples describing the given one-shot objects. As seen in the performance evaluation section, GOL can be successfully applied on various challenges, such as hand-written characters recognition or visual perception. 

The implementation of GOL for other non-image based machine learning challenges is taken into account. One such application in the field of autonomous driving is learning the model of the human driver in order to tailor the behavior of the car to the preferences of the driver. The driver model would in this case include knowledge ranging from in-car parameters settings, such as the inner temperature, to how the driver likes to drive his/her's car when the autonomous driving functions are off (e.g. how fast he/she is going, the driver likes to overtake other cars fast? etc.) 

%\nocite{*}
\bibliographystyle{IEEEtran}
\bibliography{references}

\end{document}